\documentclass[letterpaper, 10pt, conference]{ieeeconf}      

\IEEEoverridecommandlockouts                              

\usepackage{graphics} 
\usepackage{epsfig} 
\usepackage{mathptmx} 
\usepackage{times} 
\usepackage{amsmath} 
\usepackage{amssymb}  
\usepackage{siunitx}
\usepackage{hyperref}
\usepackage{algorithm,algpseudocode}

\if CLASSOPTIONcompsoc
\usepackage[caption=false,font=normalsize,labelfont=sf,textfont=sf]{subfig}
\else
\usepackage[caption=false,font=footnotesize]{subfig}
\fi

\usepackage{cleveref}

\title{\LARGE \bf
History-aware Autonomous Exploration \\
in Confined Environments using MAVs
}

\author{Christian Witting$^{1}$, Marius Fehr$^{2}$, Rik B{\"a}hnemann$^{2}$, Helen Oleynikova$^{2}$, and Roland Siegwart$^{2}$
\thanks{
$^{1}$C. Witting is a Master student at the Faculty of Electrical Engineering, Technical University of Denmark, Lyngby, Denmark.
{\tt\small witting.chr@gmail.com}
$^{2}$M. Fehr, R. B{\"a}̈hnemann, H. Oleynikova, R. Siegwart are with the Autonomous Systems Lab (ASL), ETH Z{\"u}rich, Z{\"u}rich, Switzerland. 
{\tt\small mfehr, brik, oelena, rsiegwart}@ethz.ch
This research was supported in part by the European Community’s Seventh Framework Programme (grant number n.608849).
}}%

\begin{document}

\maketitle
\thispagestyle{empty}
\pagestyle{empty}

\begin{abstract}
Many scenarios require a robot to be able to explore its 3D environment online without human supervision. This is especially relevant for inspection tasks and search and rescue missions. To solve this high-dimensional path planning problem, sampling-based exploration algorithms have proven successful. However, these do not necessarily scale well to larger environments or spaces with narrow openings.
This paper presents a 3D exploration planner based on the principles of Next-Best Views (NBVs). In this approach, a Micro-Aerial Vehicle (MAV) equipped with a limited field-of-view depth sensor randomly samples its configuration space to find promising future viewpoints. In order to obtain high sampling efficiency, our planner maintains and uses a history of visited places, and locally optimizes the robot's orientation with respect to unobserved space.
We evaluate our method in several simulated scenarios, and compare it against a state-of-the-art exploration algorithm. The experiments show substantial improvements in exploration time ($2\times$ faster), computation time, and path length, and advantages in handling difficult situations such as escaping dead-ends (up to $20\times$ faster). Finally, we validate the on-line capability of our algorithm on a computational constrained real world MAV.

\end{abstract}

\section{Introduction}
\label{sec:introduction}

Autonomous mobile robot exploration has a wide variety of applications such as visual inspection tasks, search-and-rescue missions, 3D reconstruction, or mining \cite{thrun2010a, calisi2007a, yoder2016a, rosenblatt2002a}. More specifically, Micro-Aerial Vehicles (MAVs) allow unique viewpoints and access to confined and narrow environments \cite{Zang-2016-4518}. The common goal in all exploration applications is to efficiently explore an unknown environment completely.

The research in exploration is classically divided into two fundamental approaches: the frontier-based approach and the sampling-based approach. The frontier approach seeks to maximize the map coverage by identifying and exploring the boundaries between the known and unknown parts of a map, while the sampling-based approach randomly searches the robot's configuration space for sensor poses which maximize a given objective function.

We base our algorithm on the sampling-based approach as it has been proven successful in 3D exploration and can be formulated independently of the underlying objective function, e.g., it allows multi-objective optimization of exploration, scene reconstruction and localization \cite{bircher2016a, papachristos2017a, gonza2002a}.
Unfortunately, sampling-based planner performance deteriorates in large environments or confined scenarios featuring small openings or bottle-necks. In particular tree-based exploration has the tendency to get stuck in dead-end situations where the robot needs to reevaluate and revisit already traversed sections of the map in order to find new exploration gain. Thus a large amount of computation time is wasted on searching already visited places.

In our Next-Best View Planner (NBVP) we introduce three new components to cope with the curse of dimensionality. First, our planner reduces the sampling space by locally optimizing the orientation of the sensor instead of sampling it randomly. Second, we use path simplification and smoothing methods to shorten traversal time. Third, we maintain and use a history of previously visited positions as seeds of the RRT to quickly find informative regions when the mapped area increases, and the next gain is far away.

\begin{figure}[tbp]
	\centering
	\includegraphics[width=\linewidth]{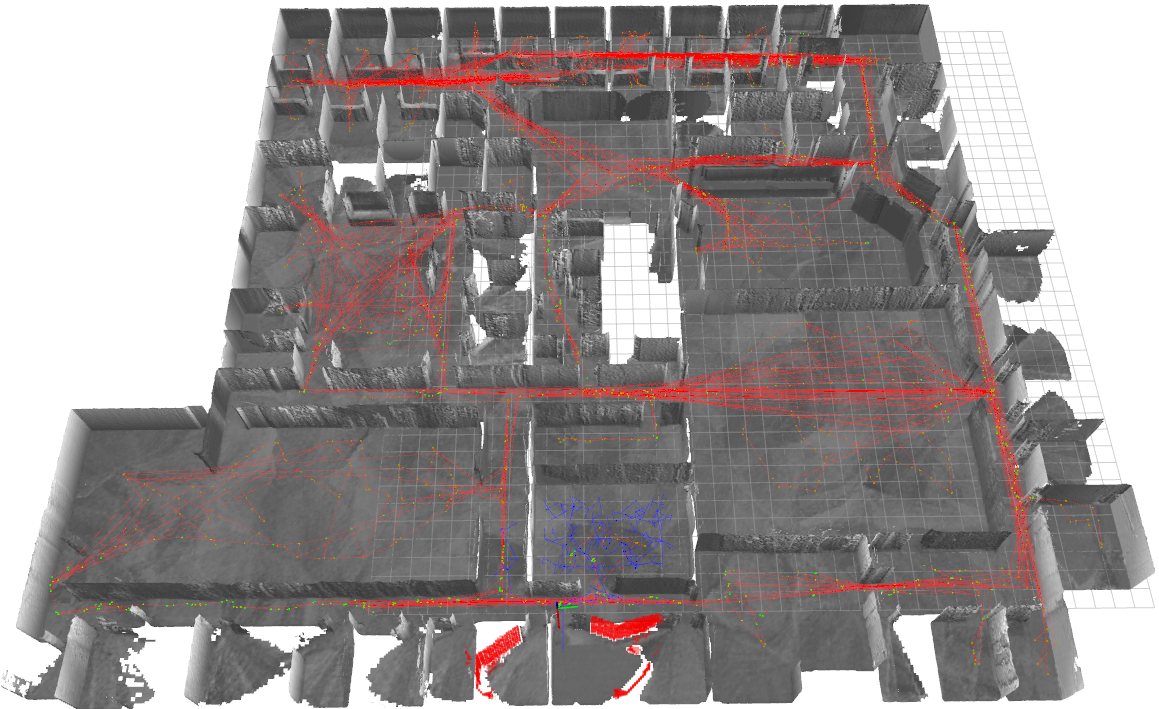}
	\caption{A map build from sensor data of a partly explored simulation of the willowgarage building. The MAV's position is the axis marker in red, green and blue, with the Rapidly-exploring Random Tree (RRT) shown in dark blue. A sparse version of the history graph is shown in red.}
	\label{fig:intro_fig}
\end{figure}

We evaluate the proposed components by comparing our planner against a state-of-the art information gain exploration algorithm, both quantitatively and qualitatively in small- and large-scale simulation scenarios. Furthermore, we validate the on-board planning capability in a real world MAV experiment.

The main contributions resulting from this work are:
\begin{itemize}
\item Boosting the RRT planning performance by using a history of exploration potential as seeds.
\item Increasing the sampling efficiency by maximizing the local sample gain w.r.t. orientation.
\item Employing dynamics-aware trajectory optimization techniques for trajectory refinement.
\item Evaluation and validation of our approach in both simulated and real world scenarios.
\end{itemize}

We organize the paper as follows. Section \ref{otherwork} presents related work. Section \ref{nbvp} introduces NBV planning and elaborates on the particular issues in exploration. In section \ref{optimized_nbvp} we present our planning algorithm. In section \ref{results} we benchmark the planner against an existing NBVP in simulation and show a real world application before we close the article with concluding remarks in section \ref{conclusion}.

\section{Related works}\label{otherwork}
Exploration planning deals with the problem of finding a set of sensor poses along the border of an unknown volume such that eventually the whole volume is explored respecting some path cost, e.g., time, length or energy. As such, the problem is closely related to the art gallery problem, the traveling salesman problem, and the problem of finding a shortest path in 3D environments, which are NP-complete individually \cite{o1983some,papadimitriou1977euclidean,canny1987new}. Additional planning constraints arise from the MAV's restricted computational and battery resources and its limitations in perception and actuation. The two most established heuristics in MAV exploration planning are frontier-based approaches and sampling-based information gathering approaches.

\subsection{Frontier-based}
The frontier-based approach is the classical approach to the exploration problem originally introduced in \cite{frontier1997a} who also extended it to multiple robots \cite{yamauchi1998a}.
In a partly explored environment there exists a boundary between known and unknown free space which denotes the frontier. Frontier-based methods extract this boundary from a map and plan a path which visits the nearest boundary.

In \cite{holz2010a} and \cite{julia2012a} the frontier method has been compared to several different exploration algorithms, and \cite{fraundorfer2012a} proposes a system which applies frontier exploration to MAVs. However, the MAV is kept at an constant height, and the comparisons are mainly made for the 2D case, whereas our method explores freely in 3D.

Traditionally, the frontier-based methods pick the closest frontier \cite{frontier1997a}. \cite{titus2017a} takes the full 3D movement of the MAV into account when extracting the route to the next frontier. Their planner does not choose the closest frontier, but the one which requires the least change in velocity within the current field of view of the robot. While this approach also achieves faster exploration rates than the NBVP we are comparing against \cite{bircher2016a}, our planner follows the sampling-based approach and thus potentially allows different exploration objectives.

\subsection{Sampling-based}
The opposing approach is sampling-based information gathering. Here the fundamental idea is to sample viewpoints in the explored map which could potentially contribute towards the exploration objective. This avoids explicit calculations of frontiers. Since the configuration space is sampled randomly, these planners also allow different optimization objectives without changing the underlying motion planning algorithm \cite{hollinger2014sampling}. \cite{bahnemann2017sampling} for example uses this approach to generate MAV system identification trajectories.

The concept of NBVs was first introduced in \cite{connolly1985a}, where the authors goal was to obtain a complete model of a scene by calculating a series of covering views. While not dealing explicitly with exploration, the idea of NBVs has been carried over into the exploration domain.

\cite{bircher2016a} uses NBVs in a 3D exploration algorithm. In a receding horizon fashion, their approach iterates between sampling accessible viewpoints in an RRT and executing the most informative path. Our algorithm builds up on theirs but introduces a memory of already visited spaces, local gain optimization, and trajectory optimization which leads to a significantly better sampling efficiency and shorter and faster explorations.

\cite{papachristos2017a} extends \cite{bircher2016a}'s NBVP to account for uncertainty in the localization and mapping between viewpoints. In \cite{gonza2002a} the uncertainty and quality of NBVs has been incorporated with respect to a simultaneous localization and mapping (SLAM) framework. In our work we assume reliable state estimation, and focus on rapid exploration.

\subsection{Other Approaches}

\cite{charrow2015a} and \cite{visser2008a} combine frontier- and sampling-based approaches. Their algorithms calculate the frontiers explicitly, but sample the poses from which frontiers can be observed. \cite{charrow2015a} also uses trajectory optimization methods to convert a piece-wise linear planned trajectory into a smooth path that obeys robot dynamics. 

\cite{shen2012a} uses particles defined by a stochastic differential equation to explore the environment. \cite{grabowski2003a} calculates an explicit region-of-interest map based on the ideas of the NBVs.

\cite{xu2017a} uses time-varying tensor fields to construct a topological skeleton of the map. Here the exploration gain is chosen based on the unknown area which can be scanned from the topological skeleton. This method resembles ours in maintaining the history of exploration potential.
However, the tensor fields does not directly translate to the 3D case needed for MAV platforms which our work is based on.

\section{Limitations of RRT-based NBV planning} \label{nbvp}
In this section we introduce sampling-based NBV planning, discuss its limitations and motivate the proposed algorithmic changes.

\subsection{Algorithm}
3D NBV planning as introduced in \cite{bircher2016a} follows a receding horizon approach.
Here the robot iterates between sampling new viewpoints uniformly in the current voxel map, and navigating towards the NBV.

In the initial step the robot grows an RRT from the current position in $(x, y, z, \theta)$-space, where $x$, $y$ and $z$ describe the position and $\theta$ describes the yaw orientation \cite{kuffner2000rrt}.
The RRT is the backbone of the planning part, and as such is responsible for finding a collision free straight-line path from the current position to a sample with gain.

For each new sample that is connected to the nearest neighbor in the tree the algorithm calculates the expected information gain.
In \cite{bircher2016a}'s NBVP, the immediate information gain is calculated by counting the number of visible unknown voxels in the sampled camera frustum exponentially discounted by its shortest distance to the current MAV position.
The total gain of a node is the summation of all immediate gains along the RRT branch to the node.

This approach of calculating samples and adding them to the tree is repeated for a predefined number of iterations or until a sample with sufficient gain is found.

Once a potential gain is found, the MAV navigates edge by edge along the shortest path found in the RRT towards the node with the highest exploration gain.
In the mean time a new RRT iteration begins, discarding all previous samples but the best branch.
When a better branch is found the robot will update its goal pose.

\subsection{Drawbacks}\label{sec:drawbacks}
\begin{figure} 
    \centering
  \subfloat[]{%
       \includegraphics[width=0.44\linewidth]{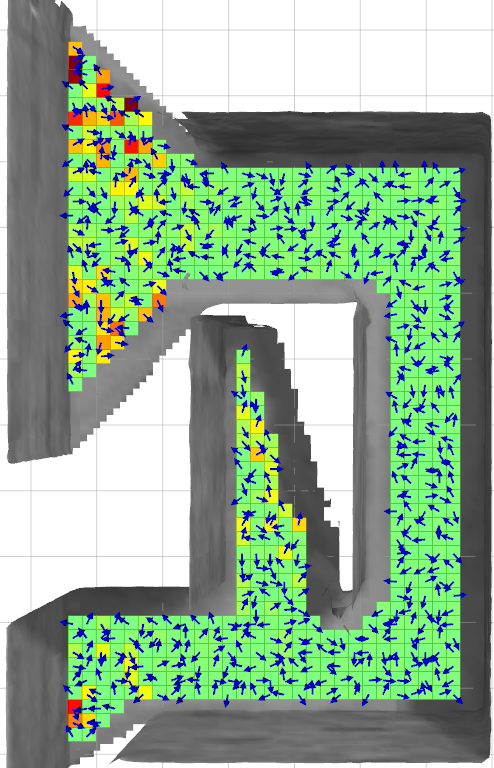}
    \label{fig:gain_random}}\hfill
  \subfloat[]{%
        \includegraphics[width=0.52\linewidth]{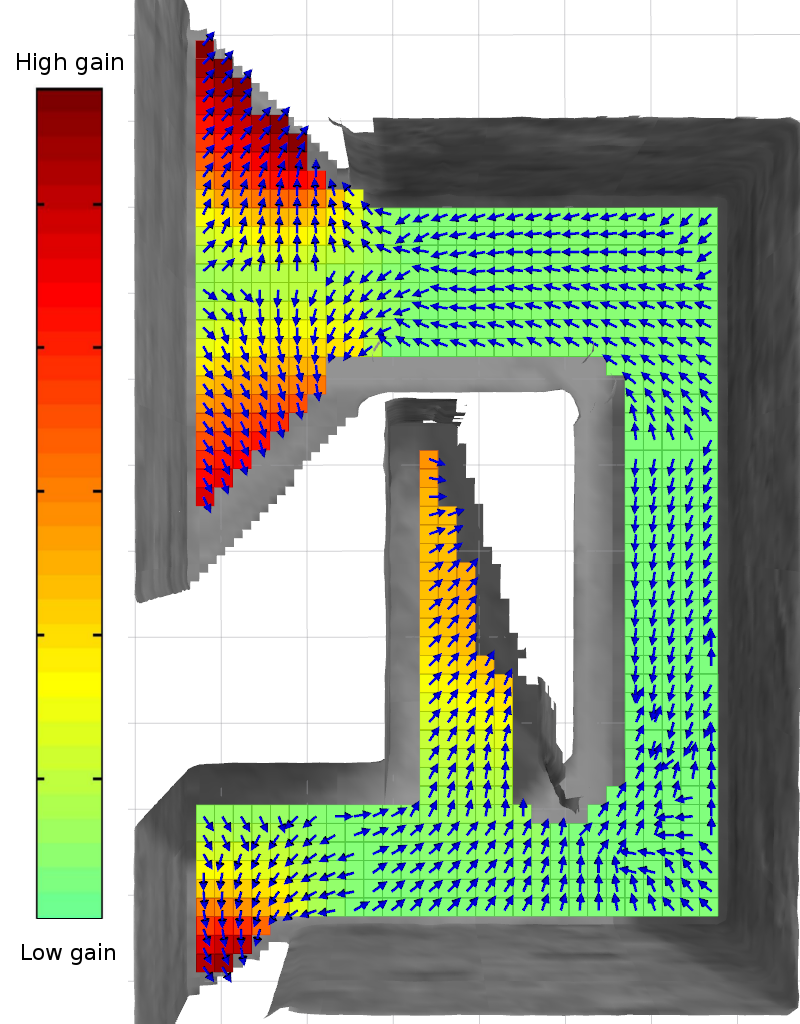}
    \label{fig:gain_opti}}
  \caption{The expected gain of a grid of poses on a slice of a map using (a) random sampling of the yaw, and (b) yaw optimization. The blue arrows represent the orientation.}
  \label{fig:comp_gain} 
\end{figure}

One of the limitations of the traditional NBVP for MAVs is the fact that the sampling takes place in the $(x, y, z, \theta)$ configuration space. 
While the sampling in $x$, $y$, $z$ is obviously desired in order for the RRT algorithm to propagate through the mapped volume, the sampling of the yaw component limits the sample efficiency of the exploration. 
For a given sampled position near unobserved voxels it is unlikely that the planner will also sample a yaw orientation that faces the camera in the optimal direction into the unobserved region and thus creates a high gain.
This is evident in Figure \ref{fig:gain_random} which visualizes the expected gain when sampling the yaw randomly.

Additionally, RRTs have a bias toward large Voronoi regions\footnote{See  \url{http://msl.cs.uiuc.edu/rrt/gallery_2drrt.html}.}.
Thus, they are sample efficient to span the Euclidean space $\mathbb{R}^3$ but take long to locally refine which is necessary to obtain different sensor viewpoints in the vicinity of the robot.

\begin{figure*} 
    \centering
  \subfloat[]{%
       \includegraphics[width=0.3\linewidth]{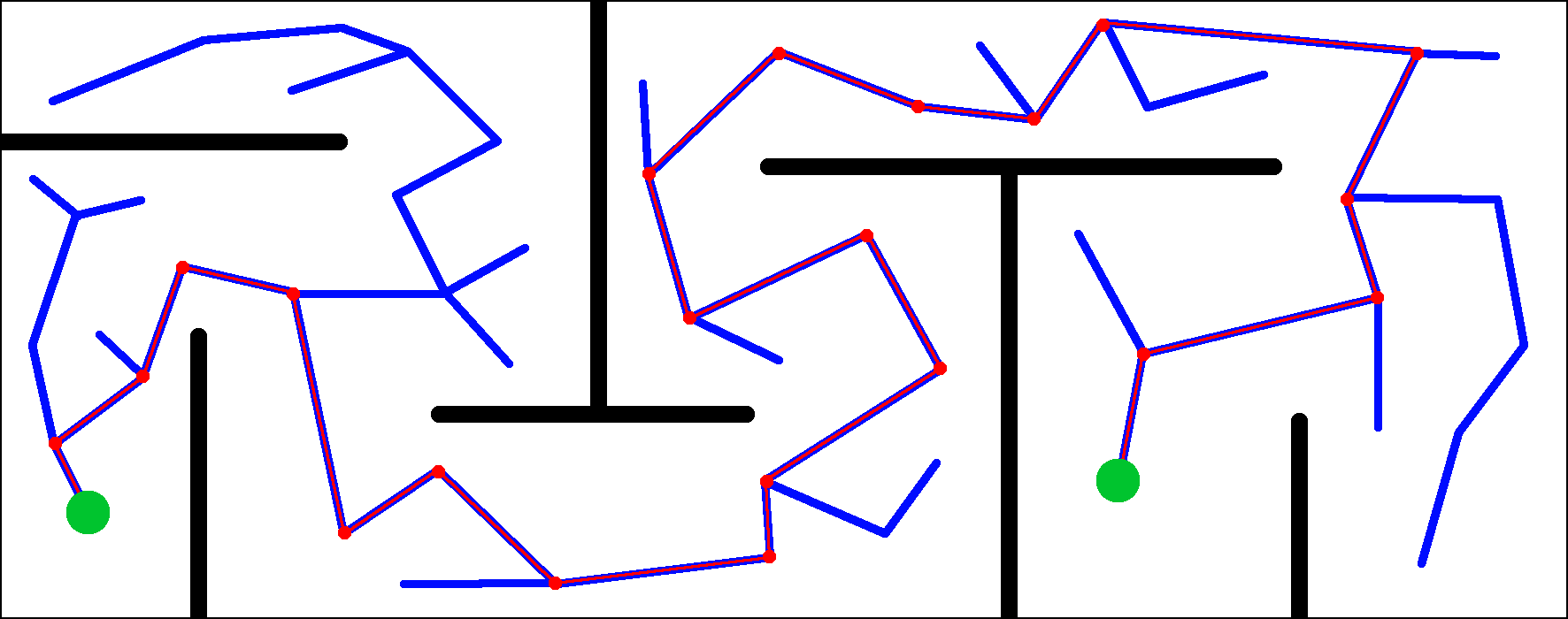}
    \label{fig:branch_bad}}
    \hfill
  \subfloat[]{%
        \includegraphics[width=0.3\linewidth]{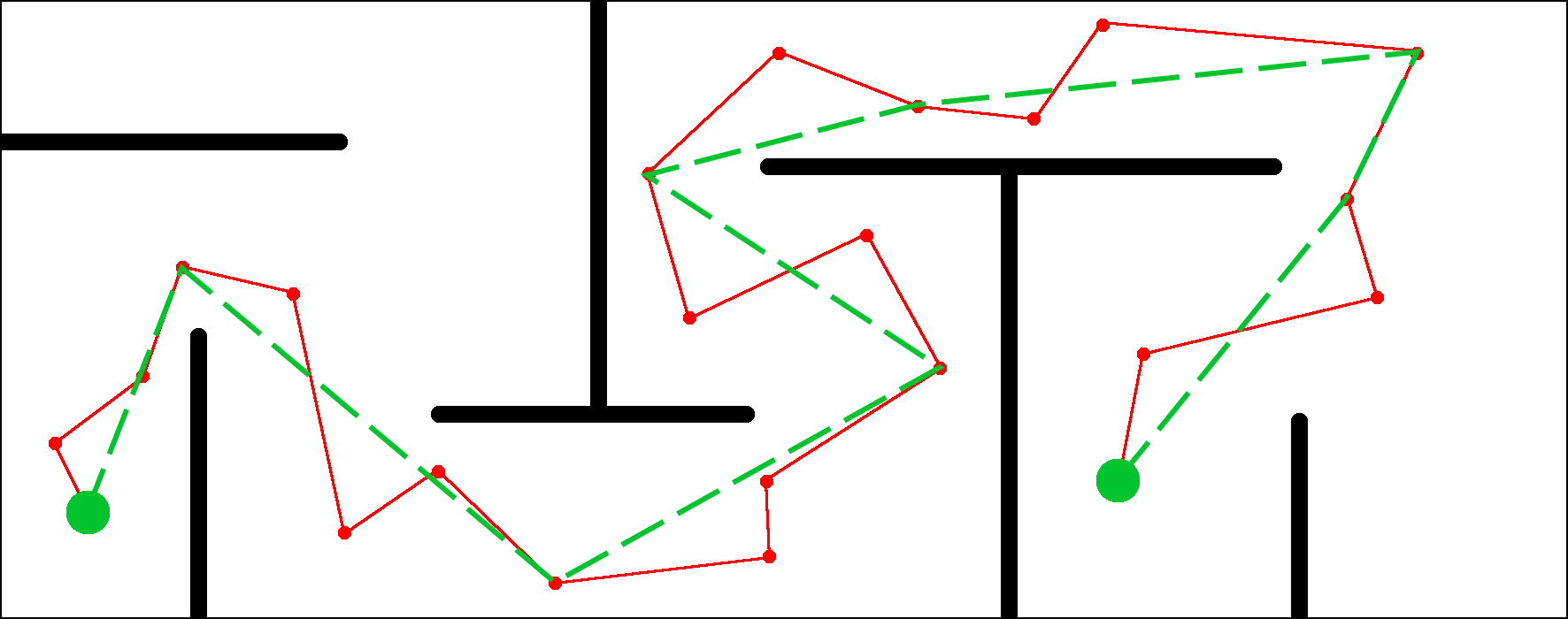}
    \label{fig:branch_better}}
    \hfill
  \subfloat[]{%
        \includegraphics[width=0.3\linewidth]{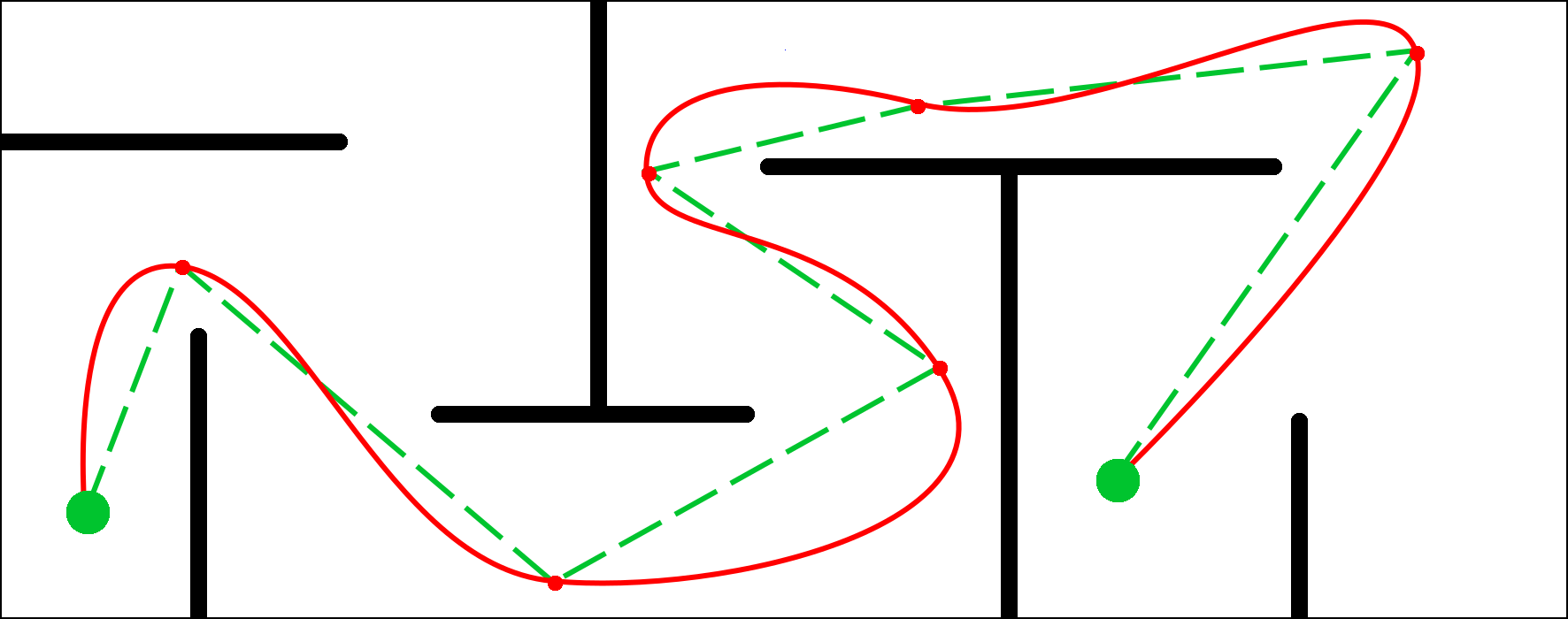}
    \label{fig:branch_best}}
  \caption{Illustration of the trajectory smoothing stages. (a) Shows the RRT tree (blue) between the start and end node (green) together with the extracted raw branch (red). (b) Shows the minimum waypoint branch (green), and (c) shows the optimized polynomial trajectory in red. }
  \label{fig:branch_bad_better_best} 
\end{figure*}

A second limitation of \cite{bircher2016a} is that the extracted waypoint list from the RRT is used directly as trajectory. 
However, this has the effect that the resulting movement is jagged due to the randomness of the RRT (illustrated in Figure \ref{fig:branch_bad}). 
Which in turn results in both long routes, and high start-stop energy consumption.

Last but not least, the RRT is a tree-based planning structure that always has its root in the current position of the MAV.
This results in a behavior where the RRT needs to be discarded and recalculated after finishing a branch.
As time passes and the mapped area increases, the distances to the next gain and thus the sampling time increases significantly.
This issue is evident in Figure \ref{fig:deadend_old} where the next gain is far away from the current position and thus the RRT tree grows immensely large.

\section{The augmented NBVP} \label{optimized_nbvp}

Based on the limitations identified in Section \ref{sec:drawbacks} we propose an augmented sampling-based NBVP.
The main intuition behind our changes compared to \cite{bircher2016a} is to reduce the sampling space to interesting areas such that the RRT can quickly find a NBV.
Our NBVP has a three-step sampling approach, a deterministic yaw policy, it generates simplified and smooth shortest path trajectories and maintains a graph of potential RRT seeds in free space as a hot-start in dead-end situations.

The exploration algorithm is described in Algorithm~\ref{alg:explore}.
In the initial step the algorithm samples a random tree with potential NBVs within the free space of a user-defined vicinity of the MAV.
Each new node gets assigned an exploration gain based solely on the number of unobserved voxels in the expected sensor frustum.
If no view with a certain gain is found in the vicinity, the robot may be stuck in a dead-end and will reseed the RRT to the closest node with a potential in the history graph. If the algorithm is still unable to find a gain within the vicinity of the new seed, the sampling will extend to the full free workspace until a sample with gain is found.

\begin{algorithm}
\caption{Exploration}\label{alg:explore}
\begin{algorithmic}[1]
\While{Map is not explored}
    \State Set sampling bounds to root vicinity
    \While{No gain found in RRT}
        \If{If initial sampling fails}
            \State Seed RRT with closest node in history
            \State Update sampling bounds
        \EndIf
        \If{Reseeded sampling fails}
            \State Increase sampling bounds to full free space
        \EndIf
        \State Sample within bounds
        \State Grow RRT
    \EndWhile
    \State Extract best branch from RRT
    \State Simplify and smooth trajectory
    \State Carry out trajectory
\EndWhile
\end{algorithmic}
\end{algorithm}

\begin{figure}[htbp]
	\centering
	\includegraphics[width=0.95\linewidth]{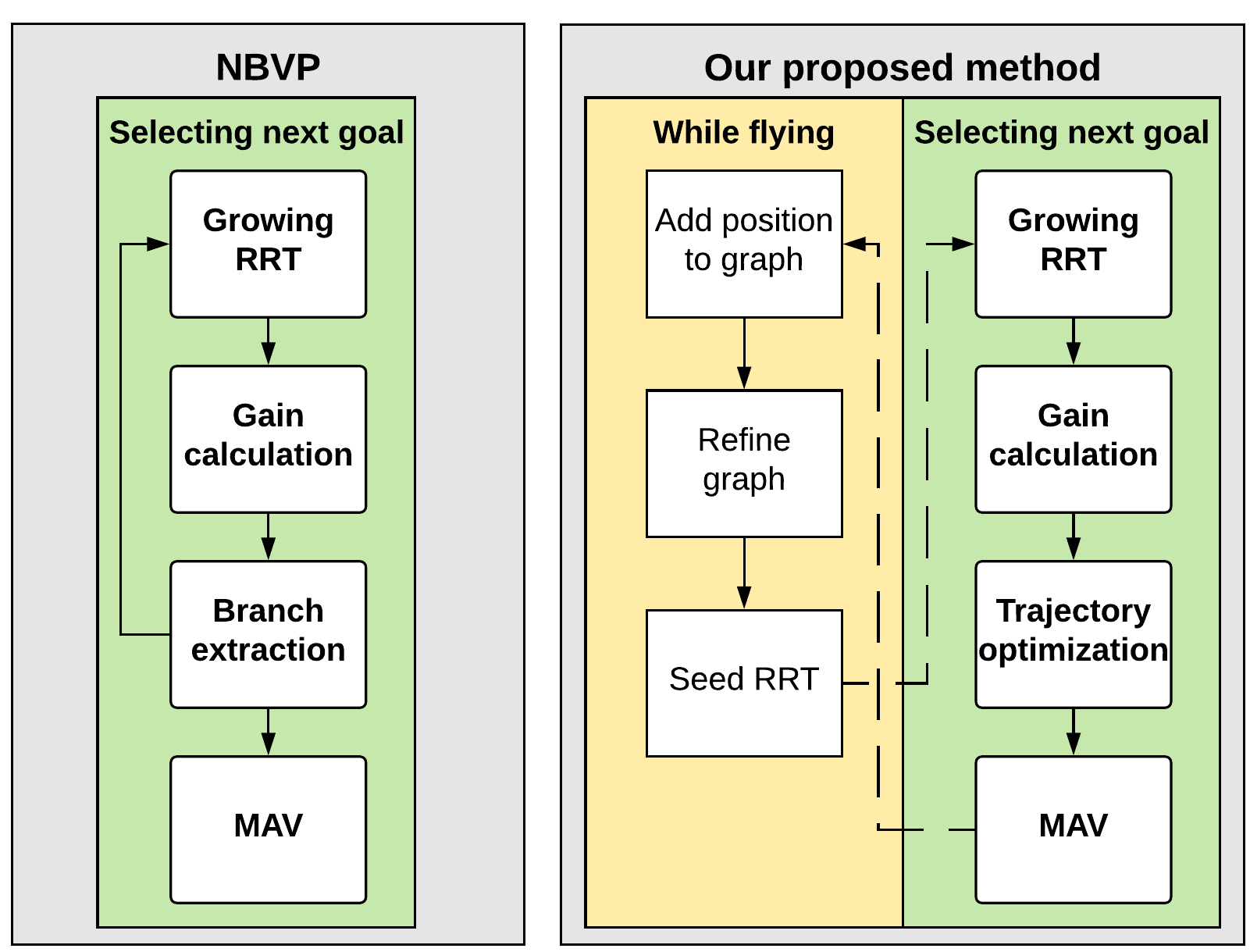}
	\caption{A comparison of the system structure of the traditional NBVP \cite{bircher2016a} and our proposed method.}
	\label{fig:system}
\end{figure}

Once a viewpoint with sufficient gain is found the MAV computes a smoothed trajectory to the node and repeats the sampling approach.
Concurrent to the exploration, the robot updates the voxel map and maintains a potential seed graph.
Figure~\ref{fig:system} shows the system in comparison to the receding horizon based planner \cite{bircher2016a}.

The mapping framework used for the exploration algorithm is based on Voxblox~\cite{voxblox2017a}. Voxblox takes a planning-centric approach to dense mapping, by maintaining both a Truncated Signed Distance Field (TSDF) and Euclidean Signed Distance Field (ESDF).
We use this representation to perform ray casting in the frustum optimization, collision checking in trajectory optimization and to refine the history graph.

The three-step sampling approach increases the chances of finding a close NBV and escape dead-ends early.
In the following we will elaborate on the extensions in orientation optimization, trajectory smoothing, and history maintainance.

\subsection{Local gain optimization}\label{sec:gain_theory}
Instead of sampling in $(x, y, z, \theta)$-space, we propose to sample only in Euclidean $(x, y, z)$-space and set $\theta$ deterministically to the optimal direction for each sample.
The optimal yaw direction is found by ray casting the sensor frustum in the set of $N$ discrete orientations $\theta \in \left\{-\pi, -\pi + \Delta \pi, \ldots, \pi - \Delta \pi \right\}$, where the discrete step size should ideally be set to $\Delta\theta = \frac{r}{R}$ to ensure that all voxels within the view frustum are considered. However, to balance the computational complexity with accuracy this value was increased to $\SI{5}{\degree}$ in the experiments here.
Here $r$ is the voxel size of the map in meters, and $R$ is the desired projection range in meters. 

We approximate the viewing frustum as a circular sector of a cylinder, with radius $R$, height $H = 2 \, R \, \sin\left(\frac{v_{\mathrm{fov}}}{2}\right)$ and central angle $h_{\mathrm{fov}}$, where $v_{fov}$ and $h_{fov}$ are the vertical and horizontal field of view of the camera respectively.

To compute the exploration gain for all $N$ frusta, we first precompute the gain of $N$ vertical slices over the whole cylinder originating from each sensor pose.
For each of these slices we cast rays from the sensor origin to the voxels on the vertical boundary line of the cylinder. 
The gain of the slice is the summation of the number of unknown voxels along the rays.
If a ray intersects with an obstacle in the map it is stopped short, in order to not count occluded unknown areas behind obstacles.

The total gain for each direction is found by summing the slice gains over a window with angular width $h_{fov}$. 
The direction with the largest gain is then the optimal direction to face for that specific sample.

Figure \ref{fig:comp_gain} shows a visual comparison of the expected gain between randomly sampled orientations and optimized orientations. 
As mentioned in Section \ref{sec:drawbacks} it is obvious from Figure \ref{fig:gain_random} that the random sampling misses a lot of the gain in several cases leading to degraded performance. 
Furthermore, it can be seen from Figure \ref{fig:gain_opti} that the yaw optimization results in finding viewpoints that consistently point towards the frontiers of the map to achieve map coverage.

\subsection{Solution selection and trajectory generation}
In the original NBVP, \cite{bircher2016a}, the MAV chooses a straight-line path according to the distance and immediate gains along the branch of the tree (see Figure \ref{fig:branch_bad}).
In our solution, we do not consider intermediate gain or distance along the path towards the NBV. 
Our planner directly navigates to the first viewpoint with sufficient exploration gain.
Since we locally optimize the orientation, the first informative sampled pose is expected to automatically be the closest gain to the seed of the RRT.
This simplification allows discarding intermediate waypoints and performing trajectory optimization.

The first step takes the jagged branch and converts it into a minimum straight-line trajectory.
The sparse waypoint list is then interpolated by continuous polyonomial trajectories as presented in \cite{richter2016polynomial} and implemented in \cite{burri2015real}\footnote{See \url{github.com/ethz-asl/mav_trajectory_generation}.}.
This results in short, smooth, and dynamically feasible trajectories as illustrated in Figure \ref{fig:branch_bad_better_best}.

\subsection{History maintenance}\label{sec:history}
As mentioned in the drawbacks in Section~\ref{sec:drawbacks}, the RRT exploration degrades in large and confined areas when the robot has to find its way towards distant information sources.
In this section we specifically address this issue by introducing a graph in free space that stores knowledge about the spatial distribution of information in already explored parts of the map.
The nodes of the graph are used to reseed the RRT when no gain is found in the vicinity of the current root.
Algorithm~\ref{alg:history} shows the full history graph algorithm that runs concurrently to the exploration algorithm.

The graph nodes consist of positions in free space that are sampled along the travelled path.
For every node, we store a measure of the exploration potential while the MAV is navigating.
This potential is a representation of exploration gain which is nearby, but not necessarily in view, and is specific to the chosen exploration objective.

To update a node, we perform a Breadth First Search (BFS) over the free voxels in its vicinity to count the number of voxel with potential gain (in our case, we count frontier voxels). 
Here the BFS ensures that there exist a collision-free but not necessarily direct connection to the gain.
By doing this for all nodes in the history graph, the potential for all the previous positions is kept up to date. 

To mitigate the effects of changes in the explored map, we attempt to maximize the clearance to obstacles of nodes in the history graph.
This is done by ascending the obstacle distance gradient in the Voxblox ESDF map when refining the poses.
This effectively warps the graph into the points of equal distance to the obstacles ensuring the connections remain collision free, approximating a Voronoi graph of the map.

As the exploration progresses, some of the nodes in the history graph become redundant when they collapse onto the same position during the refinement. 
These are continuously combined while keeping the connectivity during the maintenance step.

Figure~\ref{fig:large_maze} shows a complete history graph in an explored scenario.

\subsubsection{Seeding}
\begin{figure}[tb]
    \centering
       \includegraphics[width=0.9\linewidth]{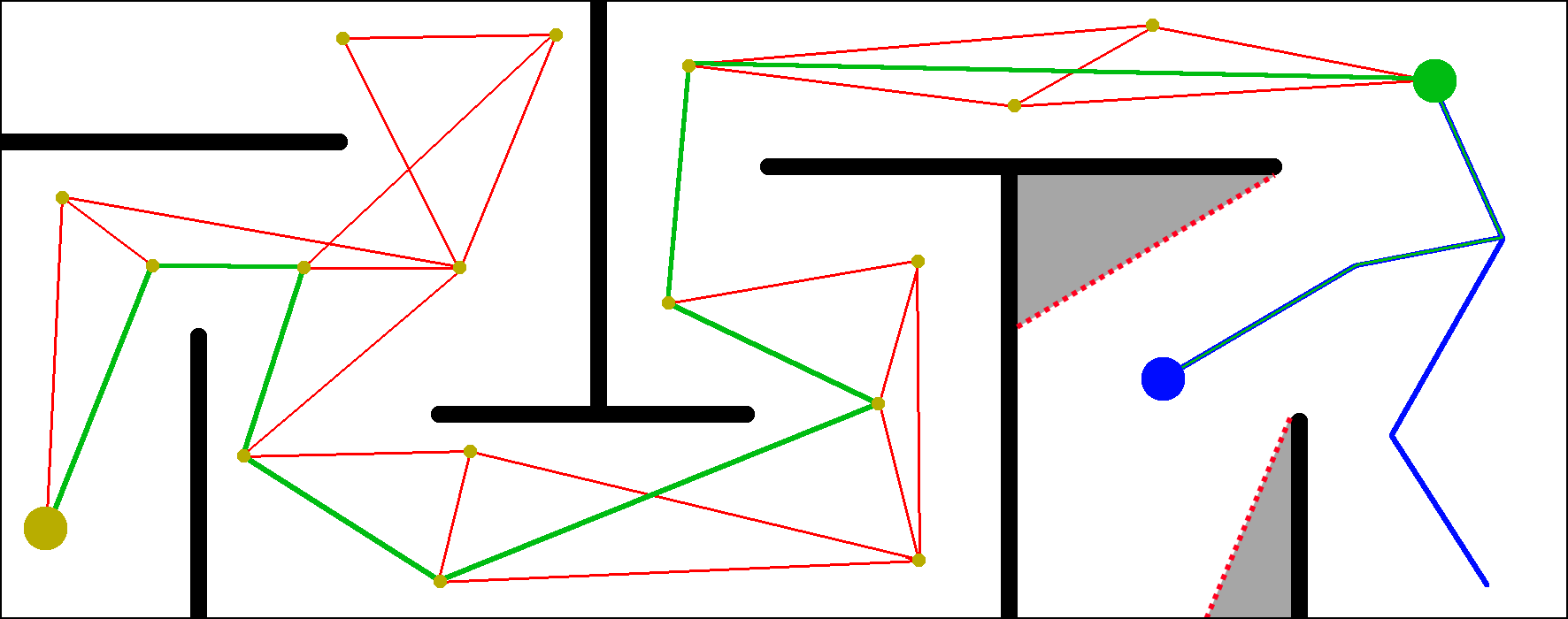}
  \caption{Illustration of the history seeding. 
  Gray denotes unknown area with the current frontier in red. 
  The large golden circle is the current position, and the red graph is the history graph. 
  The golden nodes are previous poses which do not have any exploration potential, while the green node is the closest that does. 
  It has thus been chosen as seed for the RRT tree (shown in blue). 
  The green branch is the best branch through the history graph and RRT to the large blue node which has a exploration gain.}\label{fig:graph_seed}
\end{figure}

As described in Algorithm~\ref{alg:explore} the RRT is reseeded whenever the robot cannot find informative poses in its vicinity.
In this case the closest node with potential is extracted from the history graph, and used as a seed for the RRT. 
This puts the root of the RRT in the vicinity of unexplored gain, and thus the sampling time is significantly reduced, as the RRT no longer needs to sample all the way from the current position.
The shortest path from the current position to the root of the RRT can then be extracted from the history graph.
Figure~\ref{fig:graph_seed} illustrates this process.

A comparison of a specific scenario in a dead-end can be seen on Figures \ref{fig:deadend_old} and \ref{fig:deadend_new}.

\begin{algorithm}
\caption{History maintenance}\label{alg:history}
\begin{algorithmic}[1]
\While{Robot moving}
    \State Save and connect current pose to graph
    \For{Each node in graph}
    \State Refine position of node with gradient from ESDF
    \If{Connectivity is broken}
        \State Discard refinement
    \EndIf
    \State Recalculate potential
    \If{No potential}
        \State Add to set without potential
    \EndIf
    \State Combine collapsed nodes
    \EndFor
\EndWhile
\end{algorithmic}
\end{algorithm}

\section{Results} \label{results}
We tested our algorithm both in simulation and on a real platform.
In simulation we benchmark it in a small and a large maze scenario against \cite{bircher2016a}'s sampling-based NBVP, stressing the reduced computation time, exploration time and exploration path length of our approach.
In the real-world experiment, we validate the online-capability of the planner.

A video of the system in a range of these experiments can be found on \url{https://youtu.be/Rp2bIH_e9ig}

\subsection{Small scenario}
\begin{figure}[tb]
    \centering
       \includegraphics[width=0.9\linewidth]{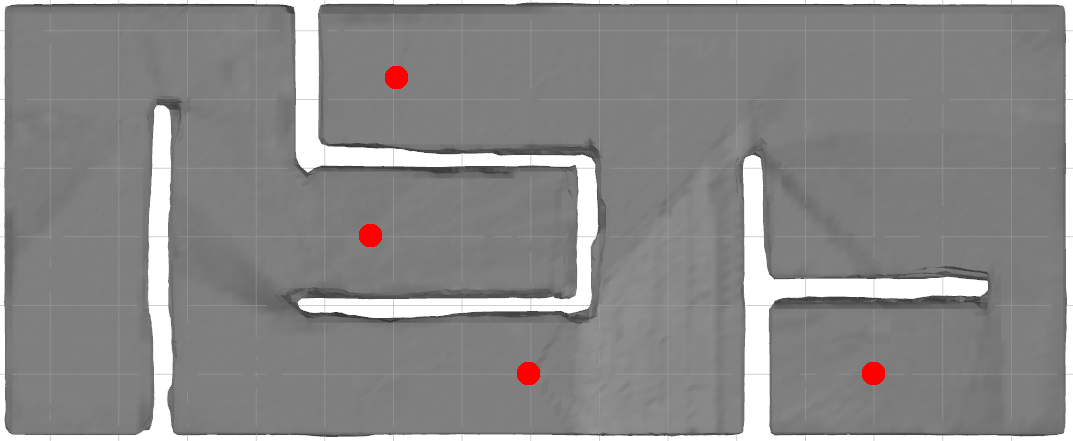}
  \caption{The map of the small test scenario ($\SI{15.5}{\meter} \, \times \, \SI{6.5}{\meter}$). The red dots are the different starting positions used for the simulations.}
  \label{fig:small_maze} 
\end{figure}

\begin{figure}[tbp]
	\centering
	\includegraphics[width=0.95\linewidth]{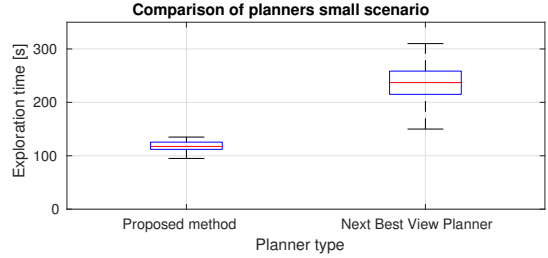}
	\caption{The boxplot for 20 exploration runs in the small scenario with our planner without history and \cite{bircher2016a} at the same maximum velocity $V_{max}=\SI{1.2}{\meter\per\second}$.
	Our improvements clearly reduces the exploration time and variance of the approach.}
	\label{fig:boxplot}
\end{figure}

We use the small maze scenario shown in Figure~\ref{fig:small_maze} to quantitatively compare the two approaches. 
As both the approaches are stochastic, they are run multiple times to account for the variance in the approach. 
In this scenario the algorithms were executed 20 times each from several different starting spots.
We omitted the reseeding in our algorithm in this scenario as it is too small to take effect.
Figure~\ref{fig:boxplot} presents the resulting exploration times.
The proposed method is on average approximately $2\times$ faster, and has a significantly smaller variance in the exploration time, showing the merits of the proposed yaw optimization, trajectory smoothing, and vicinity sampling.
Especially the yaw optimization contributes to the more deterministic behavior of the algorithm.


\subsection{Large scenario}
\begin{figure}[tbp]
	\centering
	\includegraphics[width=0.95\linewidth]{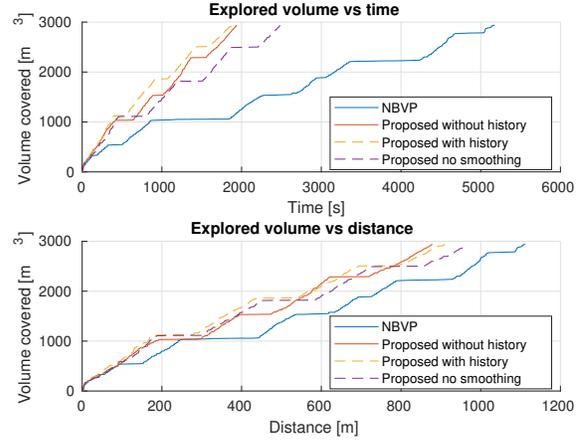}
	\caption{A comparisons with the NBVP and the proposed method in the large scenario with $V_{max}=\SI{1.2}{\meter\per\second}$. The proposed method clearly outperforms the NBVP. Furthermore, trajectory optimization results in faster and shorter exploration.}
	\label{fig:big_maze_plot}
\end{figure}

\begin{figure}[tbp]
	\centering
	\includegraphics[width=0.9\linewidth]{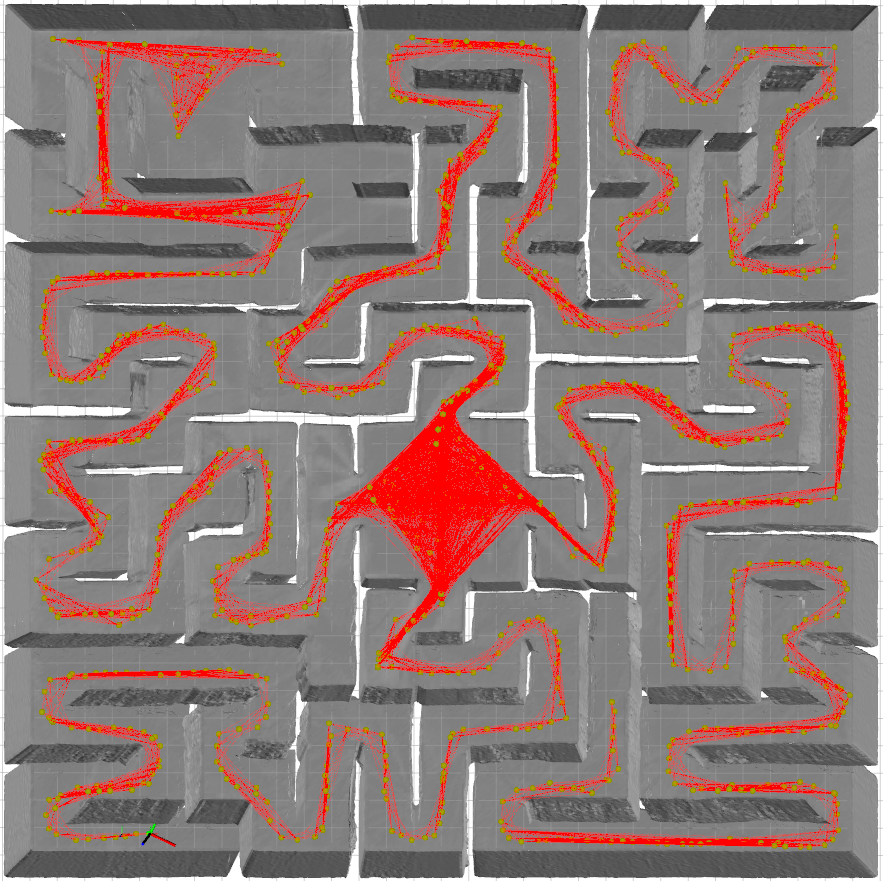}
	\caption{The fully explored map of the large scenario ($\SI{30}{\meter} \, \times \, \SI{30}{\meter}$), together with the history graph.}
	\label{fig:large_maze}
\end{figure}

In the larger maze scenario shown in Figure \ref{fig:large_maze} we compare both approaches qualitatively. 
The resulting statistics from a single experiment can be seen in Figure \ref{fig:big_maze_plot}.
It shows that the proposed method finds a solution that is both faster and shorter.
It also shows the advantage of trajectory simplification and smoothing which results in shorter trajectories with consistent velocity.

\begin{figure}[tb] 
    \centering
  \subfloat[]{%
       \includegraphics[width=0.9\linewidth]{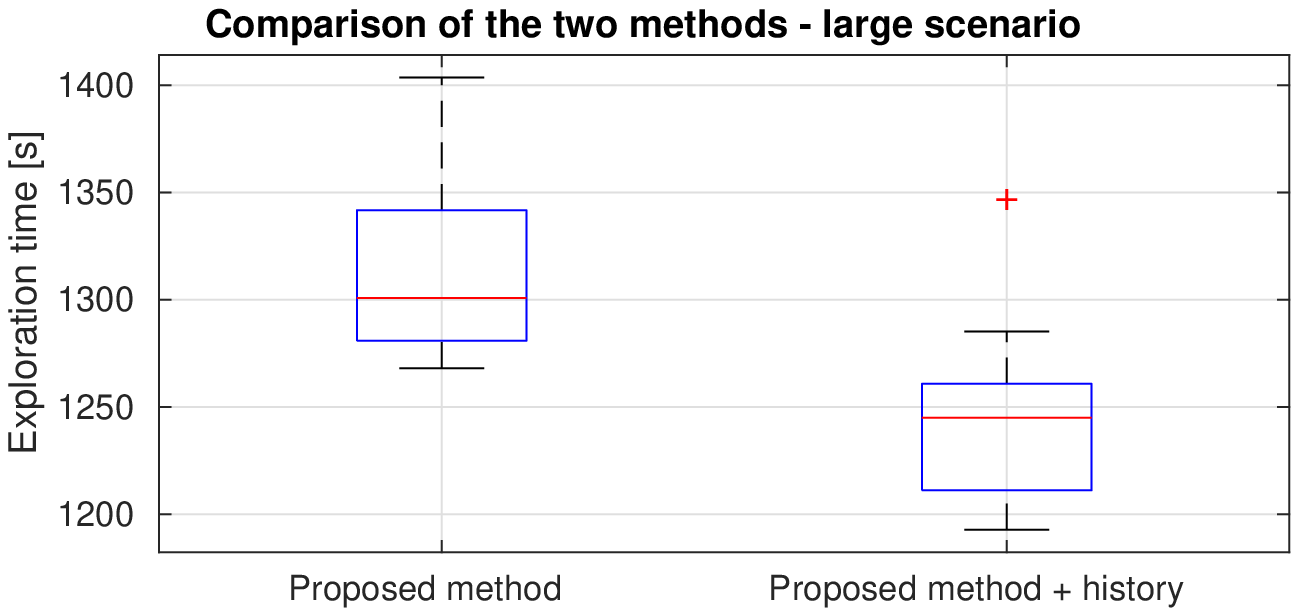}
    \label{fig:boxplot_large}}\\
  \subfloat[]{%
        \includegraphics[width=0.9\linewidth]{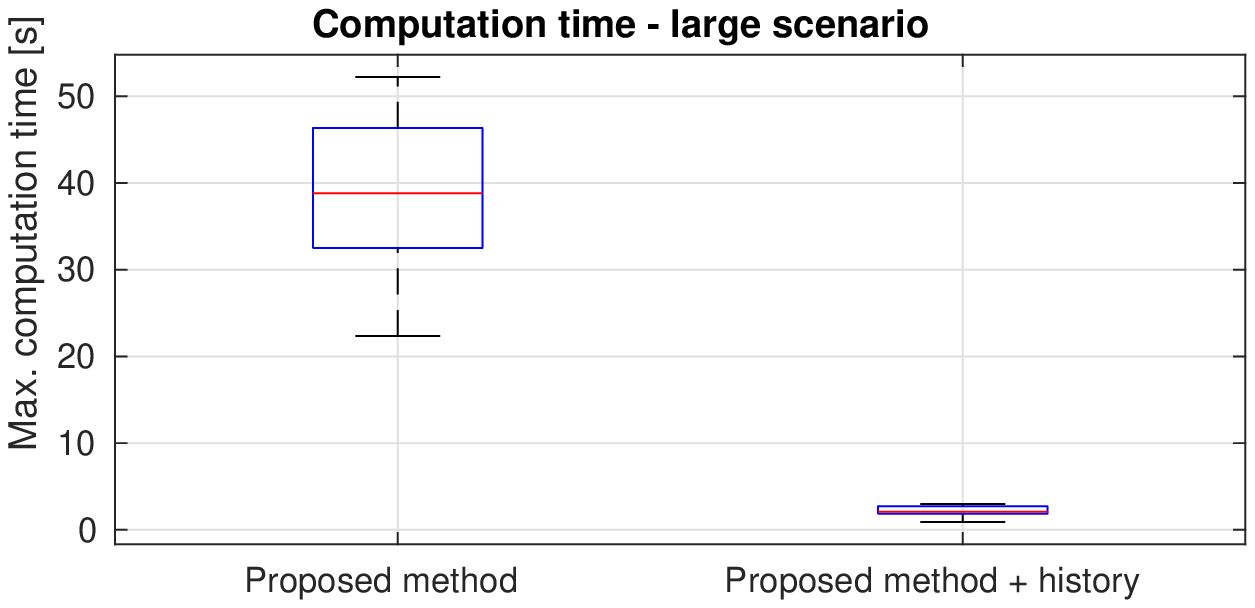}
    \label{fig:boxplot_comp}}
  \caption{The statistics for $10$ experiments for each method in the large scenario at  $V_{max}=\SI{4.5}{\meter\per\second}$. (a) shows an average reduction in exploration time of $5\%$ by using the history graph. (b) shows a reduction of the worst case computation time per iteration from $\SI{38}{\second}$ to $\SI{2.2}{\second}$ in the experiments.}
\end{figure}

A more thorough comparison of the proposed method and the history graph can be seen in Figure \ref{fig:boxplot_large}.
Here we ran the large scale experiment at higher velocities which is feasible for the controller with the trajectory optimization but infeasible with \cite{bircher2016a}'s waypoint following.
The plot shows that the history graph results in reduction of the time needed to explore the large scenario.

\subsection{Computation time}
\begin{figure}[tb]
    \centering
  \subfloat[]{%
       \includegraphics[width=0.48\linewidth]{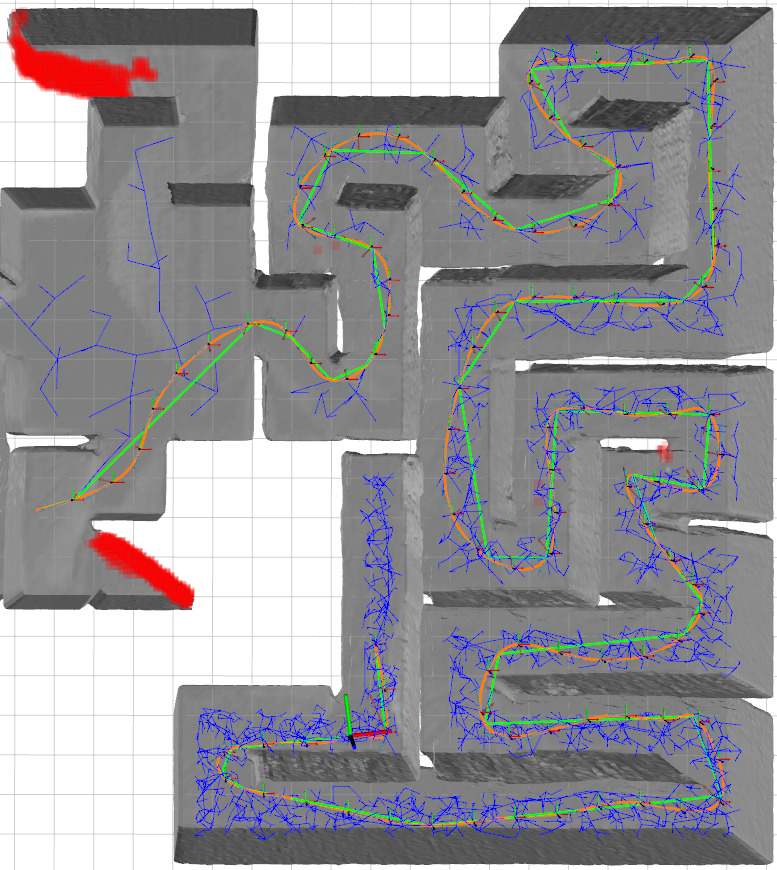}
    \label{fig:deadend_old}}\hfill
  \subfloat[]{%
        \includegraphics[width=0.48\linewidth]{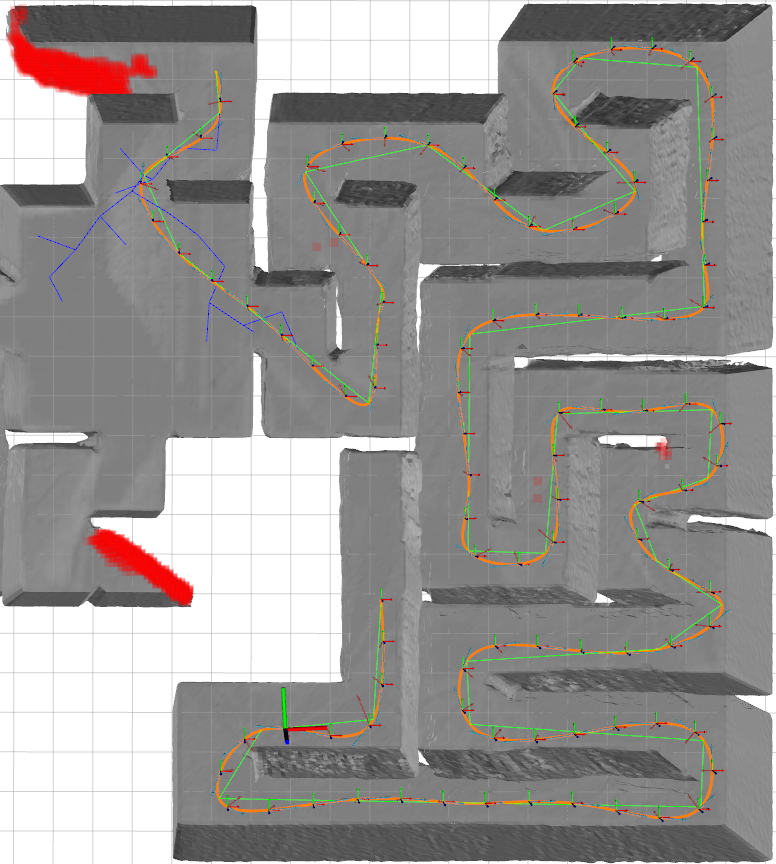}
    \label{fig:deadend_new}}
  \caption{Our method running on the large maze scenario, with a dead-end situation highlighted. Here the red part is the frontier, blue is the RRT, and green is the best branch. In (a) The tree is grown using pure RRT growing from the current position ($t_{comp} = \SI{35.3}{s}$). In (b) the RRT is reseeded using the history graph ($t_{comp} = \SI{1.4}{s}$) resulting in a much smaller tree.}
  \label{fig:deadend} 
\end{figure}

As mentioned in Section \ref{sec:drawbacks} one of the issues with the NBVP was its performance in dead-ends. 
We have specifically evaluated this scenario here in the case of the large experiment. Here the MAV's ability to escape the dead-end was evaluated. 
The result in Figure \ref{fig:deadend}, show how the seeding of the RRT in this case reduced the computation time significantly from around $\SI{35}{\second}$ to around $\SI{1.4}{\second}$. 
Furthermore, Figure \ref{fig:deadend_old} shows the excessive tree structure from the RRT without history is shown in blue.
The equivalent tree in Figure \ref{fig:deadend_new} is much smaller.

Figure \ref{fig:boxplot_comp} shows the computation time for the $10$ experiments on the large scale scenario which exploration results is in Figure \ref{fig:boxplot_large}. 
The plot shows a significant reduction of the expected maximum computation time per iteration from $\SI{38}{\second}$ to $\SI{2.2}{\second}$.

\subsection{Real world experiment}
The algorithm was also put to test in the real world. 
Here it was tested in a small room to validate that it could be run onboard a MAV. Everything was run on-board with the robot using a limited field-of-view depth sensor and VoxBlox~\cite{voxblox2017a} for mapping. 
The on-board state estimation was done with Rovio~\cite{bloesch2017a} with the recently developed localization extension Rovioli~\cite{maplab2017a}, and thus no external sensing was used.

The experiment ran in a semi-autonomous fashion. 
This was done by first letting the safety pilot map a small section of the room to give a starting point for the algorithm. 
As the room which the experiments was conducted in was open the history feature was disabled, but as there was always a potential in the MAV's current position it would not have had any effect.

On each iteration the chosen trajectory was manually approved by a human supervisor. 
This was done from a safety perspective such that the safety pilot can always operate from a safe position.

\begin{figure}[tb]
    \centering
       \includegraphics[width=0.9\linewidth]{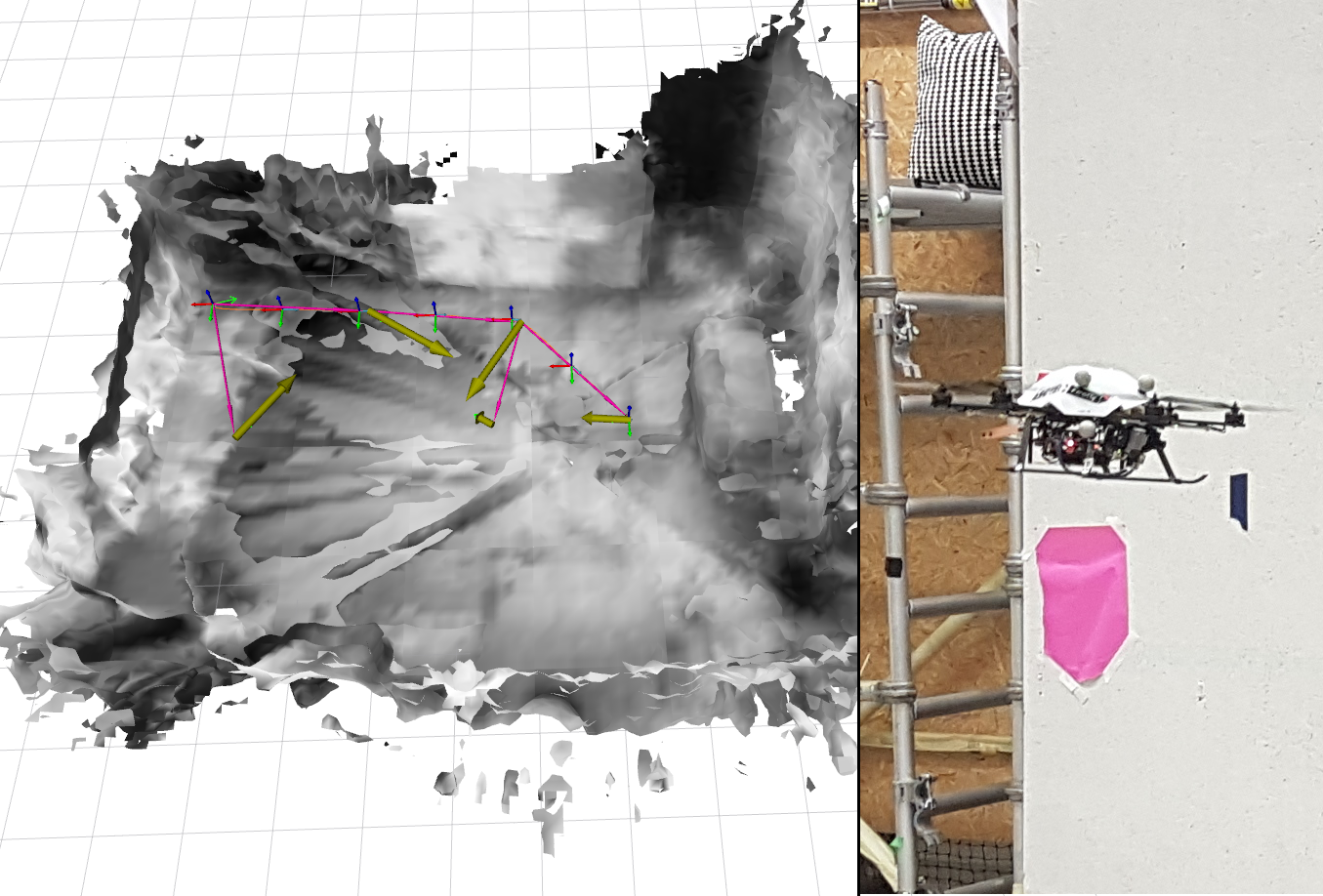}
  \caption{The map constructed in the real world scenario together with the MAV platform used for the experiment. The dimensions of the room was approximately $\SI{9}{\meter} \, \times \, \SI{6}{\meter}$.
  The pink arrows show edges in the RRT and the yellow arrows show next best sensor poses and gains.}
  \label{fig:real_map} 
\end{figure}

A figure of the constructed Voxblox map in this scenario can be seen on Figure \ref{fig:real_map}.

\section{Conclusion}\label{conclusion}
In this work we presented an improved sampling-based NBVP for autonomous exploration. Statistical simulations show a reduction in exploration time by a factor of $2$ over an existing sampling-based planner.
Our planner overcomes the curse of dimensionality of sampling-based exploration by guiding the sampling towards informative regions.
This was accomplished by introducing a history of interesting places to hot-start the exploration algorithm, which reduces the worst-case computation time to find the NBV from $\SI{38}{\second}$ to $\SI{2.2}{\second}$ in our experiments.
We formulated a geometrically optimized orientation policy which reduces the sampling space and thus the variance of the approach.
The planned trajectories were optimized to be fast, short, and dynamically feasible which allows exploration velocities of up to $V_{max}=\SI{4.5}{\meter\per\second}$ and reduces the overall exploration time.
Furthermore, we showed on-board capability in a real exploration experiment on a MAV.

\addtolength{\textheight}{-10cm}

\bibliographystyle{ieeetr} 
\bibliography{IEEEabrv,bibliography/references.bib}

\end{document}